# Building an Open Language Archives Community on the OAI Foundation

**Gary Simons**
SIL International, USA

**Steven Bird**
University of Melbourne, Australia;
University of Pennsylvania, USA

## Abstract

The Open Language Archives Community (OLAC) is an international partnership of institutions and individuals who are creating a worldwide virtual library of language resources. The Dublin Core (DC) Element Set and the OAI Protocol have provided a solid foundation for the OLAC framework. However, we need more precision in community-specific aspects of resource description than is offered by DC. Furthermore, many of the institutions and individuals who might participate in OLAC do not have the technical resources to support the OAI protocol. This paper presents our solutions to these two problems. To address the first, we have developed an extensible application profile for language resource metadata. To address the second, we have implemented Vida (the virtual data provider) and Viser (the virtual service provider), which permit community members to provide data and services without having to implement the OAI protocol. These solutions are generic and could be adopted by other specialized subcommunities.

## Introduction

Language is our principal mode of communication and our primary method for representing information. Language is also the chief embodiment of our cultural heritage and provides an important window on human cognitive ability. The list of disciplines which study some aspect of language is virtually endless: linguistics, phonetics, psychology, anthropology, philosophy, cognitive science, neuroscience, speech science, political science, history, hermeneutics, literature, language teaching, literacy, translation, dialectology, information retrieval, cryptography, and so on. The linguistic artifacts that are created and investigated by these fields include texts, transcriptions, audio and video recordings, field notes, dictionaries, grammars, and more. Frequently, a collection of such artifacts constitutes a kind of scientific database, which contains both primary observations and specialized annotations and is structured to facilitate exploration and discovery. Software technologies are used to represent this information, to model human linguistic behavior, and to automate language processing tasks.

Each of these items, whether it be data or a software tool, is a kind of language resource, and the set of disciplines involved in creating and using them is known as the language resources community. The Internet is already playing a major role for the delivery of language resources, but it is far from reaching its potential in the area of resource discovery, mainly because full-text indexing is either sub-optimal, or simply inappropriate, for many resource types. Until recently, there has been no systematic method for describing and discovering language resources.

The Open Language Archives Community (OLAC) was founded in December 2000 when a group of nearly 100 linguists, archivists, and software developers gathered in a workshop on web-based language documentation and description. After reaching consensus on the requirements for language-resource archiving (Simons and Bird, 2000a) and on a vision for how acting in community could serve to bridge the gap between the present reality and the envisioned future (Simons and Bird, 2000b), OLAC was launched with the following purpose statement:

> OLAC, the Open Language Archives Community, is an international partnership of institutions and individuals who are creating a worldwide virtual library of language resources by: (1) developing consensus on best current practice for the digital archiving of language resources, and (2) developing a network of interoperating repositories and services for housing and accessing such resources.

During its first year of operation, 2001, the basic infrastructure for OLAC was developed. During the second year, 2002, the focus has been on enlarging the community of participating archives. The standards that define the technical infrastructure have been frozen in candidate status so that member archives need not worry about a moving target as they implement an OLAC data provider. By the time of writing (late 2002), twenty-four institutions have published metadata repositories containing a total of around 30,000 records. Based on the experiences of the archives that have participated during the first two years, the standards will be refined and formally adopted by the community during the third year, 2003.

The archives currently participating in OLAC are of several types. There are conventional language archives holding physical artifacts such as documents, recordings, and images. For example, the archive of the Alaska Native Language Center has materials from the twenty languages indigenous to Alaska, collected over a period of more than 150 years. There are conventional digital archives holding large text collections. For example, the Oxford Text Archive has 2,500 texts, corpora, and reference works in over 25 different languages. There are institutions that archive the language resources collected or created by their own members. For example, LACITO (Langues et Civilisations à Tradition Orale) archives materials from Africa, Asia, and the Pacific collected over the last three decades by staff of the Centre National de la Recherche Scientifique. There are archives created by individual research projects. For example, the Comparative Bantu Online Dictionary has built a digital archive of lexicons from Bantu languages that they have used as a basis for constructing a comparative lexicographic database.

Participants like the above are archives in the traditional sense in that they hold physical and digital artifacts in order to preserve them and to facilitate access to them. Other OLAC "archives" are not so conventional; these participants are publishers and service providers who are finding that the OLAC infrastructure for resource description and discovery works well to disseminate information about their language resources to the rest of the community. For instance, the Linguistic Data Consortium (LDC) and the European Language Resources Association (ELRA) are language resource publishers; they create, collect, and disseminate digital resources that are used for developing and evaluating language technologies such as speech recognition, information extraction, machine translation, and so forth. Examples of information services are the Ethnologue, which catalogs each of the more than 7,000 living or recently extinct human languages, and the Natural Language Software Registry, which is a repository of information about language technologies.

The Open Archives Initiative (OAI), with its metadata format based on Dublin Core (DC) and its protocol for metadata harvesting, has provided a solid foundation on which to build the framework for a worldwide virtual library of language resources. But in developing that framework for the language resources community, we have encountered two main challenges. First, our community needs more precision for certain aspects of resource description than is offered by the basic DC element set [1]. Second, many of the institutions and individuals who might participate in OLAC do not have the technical resources to support the OAI protocol.

This paper presents our solutions to these two problems. The next section describes some of our needs for specialized metadata and explains our approach to implementing an extensible metadata schema and application profile. The remaining sections describe infrastructure services we have developed in order to make it easier to be a data provider or to be a service provider. Our approaches to both problems are generic and could be adopted by any specialized subcommunity.

## Specialized communities need specialized metadata

Communities are defined by their common focus. In an OAI subcommunity, such a focus is manifested in metadata. Members of a community will want to search for resources within its area of focus with high recall and precision. In order to support this, the metadata records of the community need to use specialized controlled vocabularies. Two particular qualifiers are foundational for OLAC, namely subject language and linguistic type. These are now discussed in turn.

**Subject Language.** A language resource may relate to language in general, but most typically it concerns a particular language. For instance, a given grammar may be about German. The language that a resource is *about* may not be the same as the language that a resource is *in* (e.g. the German grammar may be written in English). To express this distinction, we retain the Language element of DC and term it the *audience language.* In addition, we refine the Subject element to identify the language that the resource is *about*.

Furthermore, we need for these elements to be used with an encoding scheme for language identification in order to avoid the precision and recall problems that arise when languages are identified using character strings (for instance, "German", "Deutsch", and "Allemagne" all denote the same language, while "Low German" and "Pennsylvania German" denote languages that are different than "German"). Unfortunately, existing language identification standards, such as ISO 639 [2] and RFC 3066 [3], fall far short of meeting the needs of the language resources community, since they account for less than 10% of the world's languages and provide inadequate information about what languages the codes refer to (Constable and Simons, 2000). However, the Ethnologue (Grimes, 2000) provides a complete system of language identifiers that is well documented and openly available on the web [4].

OLAC has developed an encoding scheme (named OLAC-Language) for language identifiers. It employs the RFC 3066 extension mechanism to build a language identifier from each of the more than 7,000 codes defined in the Ethnologue. The OLAC encoding scheme also includes the 130-plus two-letter codes from ISO-639-1 that map unambiguously to a single language in the Ethnologue [5]. New identifiers for ancient and constructed languages, currently being developed by the LINGUIST List, will also be incorporated (Aristar, 2002).

**Linguistic Type.** By definition, a language resource is a resource that serves some identifiable purpose for the language community, and the discourse of the community depends on shared assumptions about these purposes. Therefore, a language resource can be categorized according to the nature of its content, viewed from a linguistic standpoint (for instance, "Is it a lexicon, a description, a primary text?"). Likewise, language software can be categorized according to the type of linguistic information it processes. As before, to improve precision and recall, we provide an encoding scheme (named OLAC-Linguistic-Type) for the type of a resource from a linguistic point of view.

Over the past two years we have developed an XML format for DC metadata, plus community-specific qualifiers, that permits OLAC metadata to be exchanged using the OAI protocol. It was implemented as a custom application of XML. The format supports element refinement using a dot notation in the tag name, and a special attribute for values from encoding schemes:

```
<subject.language code="x-sil-BAN">Dschang</subject.language>
```

The format permits free-text content to serve as an escape hatch when the coded value does not fully capture the required meaning. This also provides a migration path from legacy metadata:

(i) Dublin Core: `<subject>Language: Dschang<subject>`

(ii) Dublin Core with OLAC refinement: `<subject.language>Dschang</subject.language>`

(iii) Adding the OLAC encoding scheme: `<subject.language code="x-sil-BAN">Dschang</subject.language>`

The original OLAC metadata set included many refined elements with specialized encoding schemes in addition to those described above (Bird and Simons, 2001). We found, however, that progress on developing the needed controlled vocabularies was much slower than anticipated. Furthermore, we discovered that more specialized subcommunities within our larger community needed greater scope for experimenting with their own vocabularies. The new guidelines and format for qualified DC (Powell and Johnston, 2002; Cole *et al*, 2002) provide the extensible framework we require, and permit us to establish OLAC metadata as an application profile (Heery and Patel, 2000) built on the `dc` and `dcterms` namespace schemas.

In this new metadata model, we use the `xsi:type` attribute to represent extensions to the qualified DC metadata set. This attribute, which is built into the definition of XML Schema [6], is a directive that overrides the type definition of the current element by the type definition named in its value. For instance, the subject language example introduced above is now represented as follows:

```
<subject xsi:type="olac:language" olac:code="x-sil-BAN">Dschang</subject>
```

This simple override device allows us to specify three things: the language refinement of the subject element, the associated XML schema (via the declaration of the `olac` namespace), and a coded value for the element taken from a controlled vocabulary that is enumerated in the schema definition of the `olac:language` type. Unlike the DC approach, however, we do not put coded values (e.g., “x-sil-BAN”) in the element content. When used, element content is reserved for non-encoded, human readable values (e.g., “Dschang”) that serve as an escape hatch or that facilitate migration of legacy metadata.

All metadata extensions that are adopted by the OLAC member archives as recommended best practice for resource description are defined in the `olac` namespace schema. This new metadata format also permits individual OLAC archives and subcommunities within OLAC to set up namespace schemas that define their own refinements and encoding schemes. For example, linguists at Academia Sinica in Taipei have their own vocabulary for identifying Formosan languages, and can extend the OLAC application profile by using the `xsi:type` mechanism to incorporate their own namespace schema, as in the following example:

```
<language xsi:type="as:formosan" code="Seediq"/>
```

The standard OLAC metadata harvester harvests four things from each metadata element: the tag name, the element content, the value of the `xsi:type` attribute, and the value of the `code` attribute. Developers of third-party extensions are free to define other attributes (which could be exploited by specialized subommunity service providers); however, they are simply ignored by standard OLAC services. The appendix gives a complete OLAC metadata record that illustrates the mixture of qualified DC elements, OLAC extensions, and third-party extensions.

The OLAC protocol for metadata harvesting (Simons and Bird, 2001) builds on the OAI protocol [7]. It simply adds a few requirements to the OAI protocol. The chief among these are that the data provider must return metadata records that conform to the OLAC metadata format and that the answer to the Identify request must contain an archive description that conforms to a schema defined by OLAC. The latter gives information that an OLAC service provider may use to supply its users with a basic description of a participating archive. Elements of the archive description include the name, URL, curator, and location of the archive, the name and URL of its sponsoring institution, a synopsis of its purpose and scope, and general notes on terms of access to its collection.

## Making it easier to be a data provider

One characteristic of our subcommunity is that the institutions and individuals we want to attract as participants do not in general have the technical capability to implement an OAI data provider. Even for those that do, they may not be able to afford the investment of time. Thus it has been a priority in developing the OLAC infrastructure to make it as easy as possible to participate as a data provider. This section describes three services we have implemented in pursuit of this aim: OLACA (the OLAC Aggregator), Vida (the virtual data provider), and ORE (a forms-based repository editor).

In the OAI community the role of a metadata aggregator has come to be understood to be both a service provider and a data provider. It is a service provider in that it harvests metadata records from multiple OAI data providers; it is also a data provider in that the main service it renders is to make those records available for harvesting by others from a single source. OLACA, the OLAC Aggregator [8], provides an important service for the wider OAI community in that it provides a single source for harvesting all the metadata records that have been published by members of the OLAC community.

OLACA is also significant within the OLAC community in that it plays an important role in the strategy for making it easier to be a data provider. It does so in three ways:

1. In order to participate in the wider OAI community, a data provider must supply metadata records in the OAI_DC format. This places an extra burden on the would-be OLAC participant since the focus within our community is on using metadata extensions and specialized controlled vocabularies to give uniform and precise description for categories of particular interest to the language resources community. Supporting OAI_DC thus means implementing a conversion process. This requires not only the operations that are typical in the "dumbdown" from qualified DC to simple DC, but also the translation of coded vocabulary terms to user-friendly display labels. This is a lot to ask of each data provider. Fortunately, it is not necessary for them to do this since OLACA implements a crosswalk from OLAC metadata to the OAI_DC format. Thus, an OLAC member need only implement the OLAC metadata format, and the wider OAI community can still harvest all of its records in OAI_DC format through OLACA.
2. A second burden that individual data providers face is the cost of keeping their implementations up-to-date with the latest version of the OAI protocol. This is another way in which OLACA insulates individual OLAC members from the full cost of participating in the wider OAI community. The data provider side of OLACA keeps up-to-date with the OAI protocol, while the service provider side continues to harvest using older versions of the protocol. Thus the metadata from individual data providers who have not upgraded their implementations to the latest version of the OAI protocol are still available to the OAI community in the latest version through OLACA.

3. Looking beyond the OAI community, OLAC members also want their resources to be accessible to the web community at large. DP9 [9] is a service that makes a repository of OAI metadata accessible for discovery and indexing by conventional web crawlers. OLAC has implemented a DP9 interface to expose all the records harvested by OLACA to web crawlers [10].

Thus, in order to participate in OLAC, an institution need only implement a data provider that supports the OLAC metadata format. Given that amount of effort to participate within the community, the OLAC Aggregator takes on the burden of effort that remains for interfacing with the rest of the world by translating records to the OAI_DC format, keeping up-to-date with the latest version of the OAI protocol, and making all metadata records accessible to conventional web crawlers.

Implementing a data provider that supports the OAI protocol for metadata harvesting is not difficult for a programmer who understands CGI interfaces and dynamic database connections, but this goes beyond the capability of many institutions that could be providers of language resources. In order to make it easier for such institutions to participate, OLAC has developed a service named Vida (for "Virtual data provider"). It works in conjunction with another element of OLAC infrastructure named ORyX (for "OLAC Repository in XML"). ORyX is an XML schema that permits all of the information in a metadata repository, including identification of the archive and the content of all metadata records, to be represented in a single XML document. Vida is a process that takes an ORyX document as input and implements the OAI data provider interface as a means of accessing the information in that document. (See, for instance, the ORyX [11] that the Linguistic Data Consortium has posted to describe its collection.) Vida is a script that runs on the OLAC web site. The complete URL of the ORyX (minus the "http://") is appended to the URL for the Vida process to form the base URL for an OAI data provider that service providers can use to harvest metadata from the ORyX. See [12] for an example.

Participating in OLAC by posting an ORyX that is harvested through the virtual data provider is an ideal approach for institutions that have a relatively small collection that is already catalogued in a local database. The would-be participant need only implement a script that generates the static XML document to represent the information in its catalog database. (Later, when information in the repository needs to be updated, the script is simply run again to generate a new version of the ORyX.) The XML document is published on a publicly accessible web site, and its URL is appended to the Vida URL when registering the base URL of the new data provider. In many cases this process has taken only a few hours from start to finish. The OAI is now testing a generalized version of this approach [13].

Many potential contributors of language resources (whether they be institutions or individuals) do not have a pre-existing database describing their materials. Thus the main challenge they face before they may participate in OLAC is to create the metadata descriptions of their resources for the first time. One approach that is possible is to use the ORyX schema with an off-the-shelf XML editor, but even that represents an unacceptable learning curve for many. Thus OLAC offers a third (and still easier) way to become a data provider. This is the service named ORE, for "OLAC Repository Editor." It is a forms-based metadata editor that any potential contributor may run from a web browser. The service is hosted on behalf of the community by Linguist List [14].

The ORE home page provides a login form. It also links to a simple registration form that allows a new user to define a login password. Once logged into ORE, users may enter identification information for their archive and create metadata descriptions of archive holdings. The information is stored in a server-side database. When the contributor instructs ORE that the archive's metadata repository is ready to be published, the information in the database is written as a repository in XML. The ORyX is automatically posted at a publicly accessible URL on a central computer and registered with OLAC as a data provider serviced through Vida. Using this approach, the whole task (from writing metadata descriptions to publishing an OAI data provider) is performed start-to-finish through a forms-based application in a web browser.

**Making it easier to be a service provider**

In order for the language resources community to derive the full benefit from publishing OLAC metadata, we need for the many special-interest web sites within the community to be able to provide services based on harvested metadata records that are relevant to their special interest. For instance, a site devoted to a particular language ought to be able to offer a service that provides a catalog of all OLAC resources pertaining to that language. Or, a site that promotes XML markup of language resources ought to be able to offer a service that provides access to all known instances of language resources that have been marked up in XML. The possibilities for such special-interest services are almost endless.

Any site could, of course, construct such a service on top of a conventional OAI harvester that retrieves metadata from all the OLAC data providers. Again, however, we fear that the majority of the potential service providers within our subcommunity would not be prepared to do this. Thus another priority in developing the OLAC infrastructure has been to make it as easy as possible to participate as a service provider. This section describes two services we have implemented in pursuit of this aim: a Query function in the OLAC Aggregator, and Viser (for "Virtual service provider").

OLACA already makes it easier to be a service provider by offering a single source from which to harvest all community metadata records. But the aggregator goes one step further to facilitate special-interest service providers by adding a Query verb to the harvesting protocol. This means that services need not harvest all records, only to turn right around and discard the vast majority because they are not relevant for their service. A very practical consequence of the query facility for a narrowly focused service is that the complete set of metadata records on which the service is based will fit comfortably in a single XML response document. Thus it is not even necessary for the service provider to parse the harvested results into a database; the whole service can be built with XML tools operating on the retrieved XML document.

In addition to the six verbs of the OAI protocol for metadata harvesting, the OLAC Aggregator supports a seventh—Query. The Query verb takes an argument named *sql* to specify the selection criterion; it is expressed as the content of a WHERE clause in MySQL syntax. The selection criterion may be an arbitrary SQL expression involving tests on any number of elements from the metadata record. See the documentation of the OLACA query facility for more details and examples (Simons, 2002a). The result of an OLACA query request is an OAI ListRecords response. Like ListRecords, the Query verb uses resumption tokens to handle flow control when the query returns more hits than can be accommodated in a single response.

Using the query facility of the OLAC Aggregator directly requires that the would-be service provider write software to handle the XML response, but this goes beyond the capability of many institutions that could host special-interest service providers. In order to make it even easier to build such service providers, OLAC has developed a service named Viser (for “Virtual service provider”). It works in conjunction with the query facility of OLACA to generate a web page that displays the results of the query. For instance, a web site about the Swahili language could use a link to the following URL to generate a service provider that is a web page listing all the OLAC resources that are indexed with the language identification code for Swahili:

```
http://www.language-archives.org/tools/viser.php4?
elements=1&sql=e1.code%3D'x-sil-SWA'
&title=Swahili+Language+Resources
```

The *elements* and *sql* arguments are passed to the query facility of OLACA to generate an XML document of matching metadata records. The *title* argument is passed to a default XSL stylesheet that transforms the XML response into an HTML web page with the given title. If the XML response ends with a resumption token, the default XSL stylesheet renders it as a “More resources …” link. This link reinvokes Viser with the supplied resumption token which is in turn passed back to the query facility of OLACA. An additional argument named *xsl* is available for specifying the URL of a custom XSL stylesheet that should be used instead. See the Viser documentation for more details and examples (Simons, 2002b).

## Conclusion

This paper has described an extensible application profile for describing language resources, together with an array of tools and services that facilitate widespread participation by the language resources community. The Dublin Core Element Set and the OAI Protocol for Metadata Harvesting have provided a solid foundation on which to build this framework for the Open Language Archives Community, although both had to be extended to support needs of the community. Though the focus of OLAC is specific to its special interests, the information-sharing framework it has developed is not. It is our hope that the general solutions OLAC has found will prove helpful to other special-interest communities that want to build on the OAI foundation.

## Acknowledgments

This research has been supported by NSF Grant No. 9983258 "Linguistic Exploration" and Grant No. 9910603 "International Standards in Language Engineering." We are grateful to OLAC participants for helping to develop and test the OLAC infrastructure, and to Éva Bánik, Alan Lee, and Haejoong Lee for their work on the supporting software.

## Notes

1. http://dublincore.org/documents/1999/07/02/dces/
2. http://lcweb.loc.gov/standards/iso639-2/langhome.html
3. http://www.ietf.org/rfc/rfc3066.txt
4. http://www.ethnologue.com/codes/
5. http://www.ethnologue.com/iso639/
6. http://www.w3.org/TR/xmlschema-1/
7. http://www.openarchives.org/OAI/1.1/openarchivesprotocol.html
8. http://www.language-archives.org/cgi-bin/olaca.pl
9. http://arc.cs.odu.edu:8080/dp9/
10. http://www.language-archives.org:8082/dp9
11. http://www.ldc.upenn.edu/Projects/OLAC/corpus_table.xml
12. http://www.language-archives.org/vida/
13. http://www.openarchives.org/OAI/2.0/guidelines-static-repository.htm
14. http://www.linguistlist.org/ore/

## Appendix: Example OLAC 1.0 Metadata Record

```
<?xml version="1.0" encoding="UTF-8"?>
<olac:olac xmlns="http://purl.org/dc/elements/1.1/"
           xmlns:dcterms="http://purl.org/dc/terms/"
           xmlns:olac="http://www.language-archives.org/OLAC/1.0b1/olac.xsd"
           xmlns:software="http://www.compuling.net/projects/olac/software.xsd"
           xmlns:as-formosan="http://www.ling.sinica.edu.tw/Formosan/as-formosan.xsd"
           xmlns:netdc="http://www.ldc.upenn.edu/Projects/netdc/netdc.xsd"
           xmlns:xsi="http://www.w3.org/2001/XMLSchema-instance"
           xsi:schemaLocation="http://www.compuling.net/projects/olac/software.xsd software.xsd
                               http://www.ling.sinica.edu.tw/Formosan/as-formosan.xsd as-formosan.xsd
                               http://www.ldc.upenn.edu/Projects/netdc/netdc.xsd netdc.xsd">

<!-- OLAC extensions -->

  <subject xsi:type="olac:linguistic-field" olac:code="phonology"/>
  <contributor xsi:type="olac:role" olac:code="editor">Sapir, Edward</contributor>
  <language xsi:type="olac:language" olac:code="x-sil-BAN">Dschang</language>
  <subject xsi:type="olac:language" olac:code="x-sil-SKY"/>
  <type xsi:type="olac:linguistic-type" olac:code="lexicon">thesaurus</type>
```

```
<!-- Extensions from third-party sources -->

  <type xsi:type="software:sourcecode" olac:code="C++"/>
  <subject xsi:type="as-formosan:language" olac:code="Amis"/>
  <format xsi:type="netdc:speechformat" rate="8000" channels="2" coding="ULAW"/>

<!-- DC elements, refinements and encoding schemes -->

  <title>TITLE</title>
  <dcterms:alternative>ALTERNATIVE TITLE</dcterms:alternative>
  <date xsi:type="dcterms:W3CDTF">1963-09-14</date>
  <relation xsi:type="dcterms:URI">http://oai.grainger.uiuc.edu</relation>

</olac:olac>
```